\documentclass{article} \PassOptionsToPackage{hyphens}{url}

\PassOptionsToPackage{numbers, compress}{natbib}

 \usepackage[creativeai, final]{neurips_2025}

\usepackage{epigraph} 
\usepackage[utf8]{inputenc} 
\usepackage[T1]{fontenc}    
\usepackage{hyperref}       
\usepackage{url}            
\usepackage{booktabs}       
\usepackage{amsfonts}       
\usepackage{nicefrac}       
\usepackage{microtype}      
\usepackage[table,dvipsnames]{xcolor}         
\usepackage{tcolorbox}
\usepackage{natbib}
\usepackage[para]{footmisc}
\usepackage{caption}
\usepackage{wrapfig}
\hypersetup{
    colorlinks,
    linkcolor={red!50!black},
    citecolor={blue!50!black},
    urlcolor={blue!80!black}
}

\definecolor{set1-1}{RGB}{228,26,28}  
\definecolor{set1-2}{RGB}{55,126,184} 
\definecolor{set1-3}{RGB}{77,175,74}  
\definecolor{set1-4}{RGB}{152,78,163} 
\definecolor{set1-5}{RGB}{255,127,0}  
\definecolor{set1-6}{RGB}{255,255,51} 
\definecolor{set1-7}{RGB}{166,86,40}  
\definecolor{set1-8}{RGB}{247,129,191}
\definecolor{set1-9}{RGB}{153,153,153}

\definecolor{pastel1-1}{RGB}{251,180,174}
\definecolor{pastel1-2}{RGB}{179,205,227}
\definecolor{pastel1-3}{RGB}{204,235,197}
\definecolor{pastel1-4}{RGB}{222,203,228}
\definecolor{pastel1-5}{RGB}{254,217,166}
\definecolor{pastel1-6}{RGB}{255,255,204}
\definecolor{pastel1-7}{RGB}{229,216,189}
\definecolor{pastel1-8}{RGB}{253,218,236}
\definecolor{pastel1-9}{RGB}{242,242,242}

\definecolor{flavio}{RGB}{0, 102, 204}
\definecolor{lucas}{RGB}{0, 153, 0}
\definecolor{fred}{RGB}{204, 0, 0}
\definecolor{gio}{RGB}{128, 0, 128}
\definecolor{henrique}{RGB}{255, 102, 0}     
\definecolor{pedro}{RGB}{153, 102, 51}   

\title{Echoes of Humanity: \\ Exploring the Perceived Humanness of AI Music}

\author{%
  Flavio Figueiredo \quad
  Giovanni Martinelli \quad
  Henrique Sousa \quad
  Pedro Rodrigues \AND
  Frederico Pedrosa \quad
  Lucas N. Ferreira  \AND
  \textnormal{Universidade Federal de Minas Gerais (UFMG)}
}

\begin{document}

\maketitle

\begin{abstract}
Recent advances in AI music (AIM) generation services are currently transforming the music industry. Given these advances, understanding how humans perceive AIM is crucial both to educate users on identifying AIM songs, and, conversely, to improve current models. We present results from a listener-focused experiment aimed at understanding how humans perceive AIM. In a blind, Turing-like test, participants were asked to distinguish, from a pair, the AIM and human-made song. We contrast with other studies by utilizing a randomized controlled crossover trial that controls for pairwise similarity and allows for a causal interpretation. We are also the first study to employ a novel, author-uncontrolled dataset of AIM songs from real-world usage of commercial models (i.e., Suno). We establish that listeners' reliability in distinguishing AIM causally increases when pairs are similar. Lastly, we conduct a mixed-methods content analysis of listeners’ free-form feedback, revealing a focus on vocal and technical cues in their judgments.

\end{abstract}

\section{Introduction} \label{s:intro}


In the book {\em I, Robot} (1950), Isaac Asimov wrote a passage stating that: ``you just can't differentiate between a robot and the very best of humans.''.
This quote offers a provocative view on the evolution of AI. In a way, Asimov challenges us to consider that AI could become comparable to the very best of humanity. Over the last few decades, we've indeed seen AI outperform humans in several objective tasks with a clear score (e.g.,  image recognition or playing Go)~\cite{He2015,Russakovsky2015,Chen2024,Silver2016,Silver2018}. 
However, when we shift our focus to artistic content, the subjectivity of aesthetic assessment makes it such that we no longer have a {\em single} score of correctness~\cite{Lerch2025}. In these settings, Asimov's statement becomes even more grounded, as we shift from improvements in scores to subjective differentiation. This poses the techno-philosophical question of what differentiates authentic (human) from artificial (AI). 

From a musical perspective, this question can be rephrased as: {\em What Sounds Like an AI?} To understand its relevance, consider the recent advances in AI Music (AIM). Nowadays, services such as Suno\footnote{\url{https://suno.ai}} and Udio\footnote{\url{https://udio.com}} base their business models on generating songs from textual prompts. These services are changing several aspects of music composition~\cite{Lerch2025}, production \cite{Musictech2025}, and consumption~\cite{Musically2025}. In fact, synthetic music projects that release exclusively AIM, e.g., {\em The Velvet Sundown}~\cite{Atlantic2025}, are now reaching millions of streams. Within this context, understanding how humans perceive AIM is essential not only for educating users on how to identify AIM, but also, conversely, to improve AI models. Thus, motivated by these industry disruptive advances, we focus on investigating the hypothesis below:

{\em Human listeners rely on contextually grounded cues (e.g., as repetitive structure or synthetic-sounding vocals) that help discern whether a piece of music is AIM or human-made.}

We study this hypothesis based on a {\em listener-focused} Turing~\cite{Turing1950}-like blind test, in which listeners are exposed to a human-made song and an AIM song. While similar listening experiments may be employed to assess new AIM models~\cite{Donahue2019,Hernandez2022}, in this case, our object of study is not a model or algorithm, but the listener themself. Specifically, consider that confusion between AIM and human-made songs is commonly regarded as a positive outcome for a new model. Our study does not seek this result. Our focus lies in a deeper examination of {\em when} and {\em how} listeners differentiate. 

To tackle the {\em when}, we make use of a \textit{randomized controlled crossover trial}~\cite{Jones2003}. In this trial, some pairs of songs are highly similar. This is the treatment group. Other pairs are uniformly random pairs of songs. The control group. Listeners also evaluate multiple pairs of songs --- the longitudinal aspect --- wherein the order of pairs is randomized for each listener. The initial audio in each pair is also randomized. To address {\em how} listeners differentiate, we explore a mixed-methods~\cite{Doyle2009} grounded-theory content analysis~\cite{Corbin1990} on the text feedback provided on why they made their choices.

Unlike other studies~\cite{Grotschla2025,Sarmento2024,Hernandez2022}, and to reduce bias, we explore a novel {\em in-the-wild-style} dataset of AIM songs, sourced from users sharing AIM on Reddit\footnote{\url{https://reddit.com/r/SunoAI/}}. In this sense, AIM songs are derived from real-world usage without any prior connection to our research. Human-made songs were obtained from independent artists on Jamendo\footnote{\url{https://jamendo.com}} via the MTG-Jamendo dataset~\cite{Bogdanov2019}. This dataset was gathered in 2019 and predates the commercial rise of AIM (Suno was released in 2023). To gather a diverse set of feedback, our study sampled from two distinct listener populations: one composed of volunteers initially recruited from (but not limited to) the Computer Science and Music Departments of a major Brazilian university, and another composed of crowd-workers hired through Prolific\footnote{\url{https://prolific.com/}}.

Our study shows that {\em when pairs are random} listeners cannot differentiate AIM from human-made songs, i.e., they are no better than random guessing. Nevertheless, {\em when pairs are highly similar}, listeners are able to make this distinction. Our results also show that longer exposure to {\em practical musical experience} (e.g., playing an instrument) and prior knowledge of AIM also increase the chance of AIM identification. Our content analysis indicates that a focus on vocal and technical aspects again leads to an increases on AIM identification. These results have implications on how to make AIM sound more human-like and on how we may be used to educate users on how to identify AIM.


We now discuss related work. Section~\ref{s:mm} describes our experiment, Section~\ref{s:results} results, and Section~\ref{s:conc} conclusions. Appendices \ref{appn:ethics} to \ref{appn:coding} present an ethical statement, limitations, and complementary results.

\section{Related Work} \label{s:rw}


With the rise of generative models in recent years, the intersection of AI and art in general has attracted significant attention~\cite{Lerch2025,Cetinic2022,Epstein2023,Yang2020,Ji2023,Le2025}. As our work focuses on musical content, the discussion in this section will primarily address music-related studies. Before doing so, we note that perception studies for images have existed since at least the 1960s. In particular, the seminal work of Michael Noll~\cite{Noll1966} who closely mimicked one of Mondrian's work, {\em Composition with Lines (1917)}, using a computer. In this very particular setting, participants often could not tell what as human-made versus computer-generated. Recent studies reach similar conclusions when examining generative AI for images~\cite{Elgammal2017,Ragot2020,Xu2024} and poetry~\cite{Kobis2021}. More focused on human perception rather than accuracy, {\em as we are}, Candello et al.~\cite{Candello2017} found that humans perceive some typefaces as more machine-like.

These efforts, however, steer the participant toward focusing on particular characteristics, such as {\em beauty} or {\em novelty}. The authors also do not employ a {\em randomized controlled crossover trial} (RCCT)~\cite{Jones2003} as we do. Using the RCCT here {\em causally} shows that participants are able to make this distinction when content is closely related. We also do not steer participants towards any particular cues that may guide their perception. In our study, we let participants describe why they made their choices in free-text form, which is analyzed via a mixed-methods~\cite{Doyle2009,Corbin1990} content analysis.




Within music, Sarmento et al.~\cite{Sarmento2024} analyzed how humans perceived the quality and the authenticity of rock and progressive metal AIM. Whereas Donahue et al.~\cite{Donahue2019} and Hernandez et al.~\cite{Hernandez2022} used a similar Turing-style test to benchmark AIM models. Both studies focus on symbolic AIM generated by the authors themselves. In contrast, here we focus on AIM audio generated real-world of user interactions with commercial text-to-music models. We also point to two very recent {\em unpublished pre-prints}~\cite{Lecamwasam2025,White2025} focused on perception. These ongoing endeavors, however, either focus on particular aspects of perception such (i.e., emotions and/or expression), and again explore author created AIM.




Our research hypothesis focuses on the broader issue of unveiling the cues humans employ to determine AIM, without steering participants towards particular cues. Also, our AIM dataset is both distinct from and richer than the ones explored by prior work. We present a novel dataset of uncontrolled (songs not created by the researchers) AIM songs in {\em audio}, not symbolic form. Furthermore, our study consists of an RCCT and a mixed-methods content analysis. Neither approaches were explored by these efforts. Our research allows both for a causal exploration of our hypotheses (via the RCCT), as well as an in-depth exploration of free-text feedback (via the content analysis).


Focused on user preferences --- {\em rather than perception} --- Grötschla et al.~\cite{Grotschla2025} benchmarked commercial and open-source text-to-music models against songs from independent artists. In their study, AIM songs were generated by first extracting the descriptive textual tags associated with each song in MTG-Jamendo. Using these tags, the authors hand-curated the creation of AIM songs, forming closely related pairs. Their findings show a general participant preference for AIM songs. 

Building on this discussion, we now present our novelty statement. Different from other studies, we compare real-world human-made compositions with non-curated user-generated AIM songs. Our experiment is larger in scale than prior perception-focused studies and is the first to explore our specific hypothesis in depth using a causal, quantitative RCCT as well as a mixed-methods analysis.

\section{Materials and Methods} \label{s:mm}


To gather our dataset of AIM songs, we crawled Reddit's \texttt{r/Suno} community where users post their songs via YouTube or Suno links. Our crawler gathered posts from July 21, 2023, up to February 25, 2025. A total of 33,626 posts were gathered. Posts may be accompanied by textual tags, known as {\em flairs}, that indicate the content of the post: \texttt{Song, Song - Audio Upload, Song - Human Written Lyrics, and/or Song - Meme}. From these, we ignored the Meme songs due to their comedic and distinctive nature. From the other flairs, audios were downloaded by following Suno (4,059 songs) and/or YouTube (8,315) links. These songs were paired with ones from MTG-Jamendo.


In our study, we attempted to reduce the impact of the music genre, since AIM models can perform better or worse for different musical genres \cite{copet2023simple}.  To do so, we used the descriptive genre tags (87 in total) that accompany MTG-Jamendo. For each AIM song, we employed the {\em Essentia MTG-Jamendo} classification model~\cite{Alonso2020} and kept the genre if the classification score (confidence) was greater than $0.4$. This cut-off is much higher than a random classification of 87 genres (confidence of $0.011$) and represents the top quartile of scores.

\subsection{Creating Song Pairs}

We controlled for similarity by creating two sets of pairs: the {\em random} set and the {\em similar} set. For the {\em random} set, we uniformly at random filtered 5 songs for each dataset with the genres {\em pop, rock, hip-hop, electronic, and metal}. To reduce language biases, we manually made-sure that these songs either had no lyrics or English lyrics. Moreover, we also made sure to select songs with durations ranging from 1.5 minutes to 4 minutes. If these criteria were not met, another random song was considered until we covered one song per genre per dataset (10 in total). These songs were then used to generate uniformly at random AIM by MTG-Jamendo song pairs per participant. For each participant, we ensured that once a song was used, no other pair with that song could be created.



For the {\em similar} set, we used the songs' audio to obtain 10 similar pairs of songs. For both the human-made and AIM songs, we generated CLAP embeddings \cite{Wu2024}, using the same settings as \cite{Grotschla2025}. Next, we calculated the cosine similarity between the embeddings of all AIM and human-composed songs that shared the same musical genre. We then filtered out any pair with similarity below $0.8$ (less than 2.5\% of pairs reached this high similarity). Pairs were then ranked by similarity. We manually chose the 10 most similar pairs that: (1) had same duration criteria and lyrics criteria as in the {\em random} set; and, (2) did not share any song with a previously more similar pair. In the end, we were left with 3 pairs of \textit{electronic} music, 2 of \textit{rock}, 1 of each: \textit{classical, ambient, hip-hop, pop}, or \textit{metal} music. 







\subsection{The Humanness Perception Study (https://hpc.uai.science)}

Before starting the study, we asked participants for their email addresses. This email was stored using a one-way hash to keep track of responses. After this introduction, participants were asked to evaluate five pairs of songs. The four initial pairs were randomly generated for each participant. Two of these pairs were created from the {\em random} pool, two others came from the predefined pairs in the {\em similar} pool. These four pairs were shuffled to avoid ordering biases. For each pair and participant, the song order for that pair was also uniformly randomized. We recall that songs were never repeated across pairs. Participants were then presented with a final gold-standard (trap) fifth pair. This pair consisted of the introduction of Symphony No. 5 in C minor, Op. 67 by Ludwig van Beethoven, arguably one of the most famous musical introductions in music history. This song was paired with an AIM song created by the authors that starts with the verse: ``This is not a human song, I'll say it right away''. 

Each pair was presented on its own page. While progressing through the study, participants were not allowed to change their answers or to skip pairs. Song titles were also not shown to participants. For each pair, a web form initially asked participants to choose only one of five choices: \textit{A. The first song was generated by AI}; \textit{B. The second song was generated by AI}; \textit{C. Neither song was generated by AI}; \textit{D. Both songs were generated by AI}; and \textit{E. I cannot state if these songs were generated by AI or by humans}. For each pair, participants could also indicate whether they had already heard: {\em A. The first song}; {\em B. The second song}; {\em C. Neither song}; {\em D. Both songs}. Furthermore, for each pair, participants were also presented with an optional free-text input field to share any opinions about the songs or explain their choice. Finally, we also internally logged how long participants spent on each pair.

After rating the pairs, participants could optionally answer a survey asking for: age, native language, formal musical education, and practical musical experience (i.e., how long they had played any instrument), and whether they were familiar with AIM services/models (true or false). Experience and education were binned by: \textit{Less than one year, one to five years, five to ten years, over ten years}. 


Participants came from two distinct populations, one volunteer-based and one crowd-worker-based. The volunteer pool was initially seeded through social media posts from the Computer Science and Music Departments of a major Brazilian university. The study was also featured on the university website\footnote{\url{https://hpc.uai.science}} and in news outlets. After participating, participants were encouraged to share the study link to any contacts they wished. The crowd-worker pool consisted of 100 English-speaking participants from Prolific\footnote{Of 110 English speakers, 10 found the study via the volunteer experiment.}. These participants were informed that they could take as long as they wished in the study and were payed 2 GBPs. Observe that by exploring these two populations, our study allows both for a more diverse demographic representation, as well as access to different motivations when providing answers (extrinsic payments versus intrinsic volunteering)~\cite{Santy2025}.
Our pairs, responses, and source code are available at: \url{https://github.com/uai-ufmg/hp-study}.




\section{Results} \label{s:results}

From June 06th to July 30th 2025, 653 participants logged to our study's website. For reliability,  we consider only participants that: (1) both knew and correctly identified the Beethoven piece in the last pair (337 out of 653); and, (2) did not know {\em any} songs in the four initial pairs (308 out of 337). This leads to 1,232 answers (308 $\times$ 4). We observe that filter (1) removes any participant that did not reach the fifth pair. The rate of participants that both knew and identified the fifth pair is 66\% (the aforementioned 337 out of 504 that reached the fifth pair). We also observe that filter (2) removes any prior knowledge of songs. Given that our songs are either from smaller independent artists or AIM, the knowledge of any song raises a doubt on whether participants were paying attention. 

\begin{figure}[t!] 
    \centering
    \begin{minipage}{0.48\textwidth}
        \centering
        \includegraphics[width=0.80\linewidth]{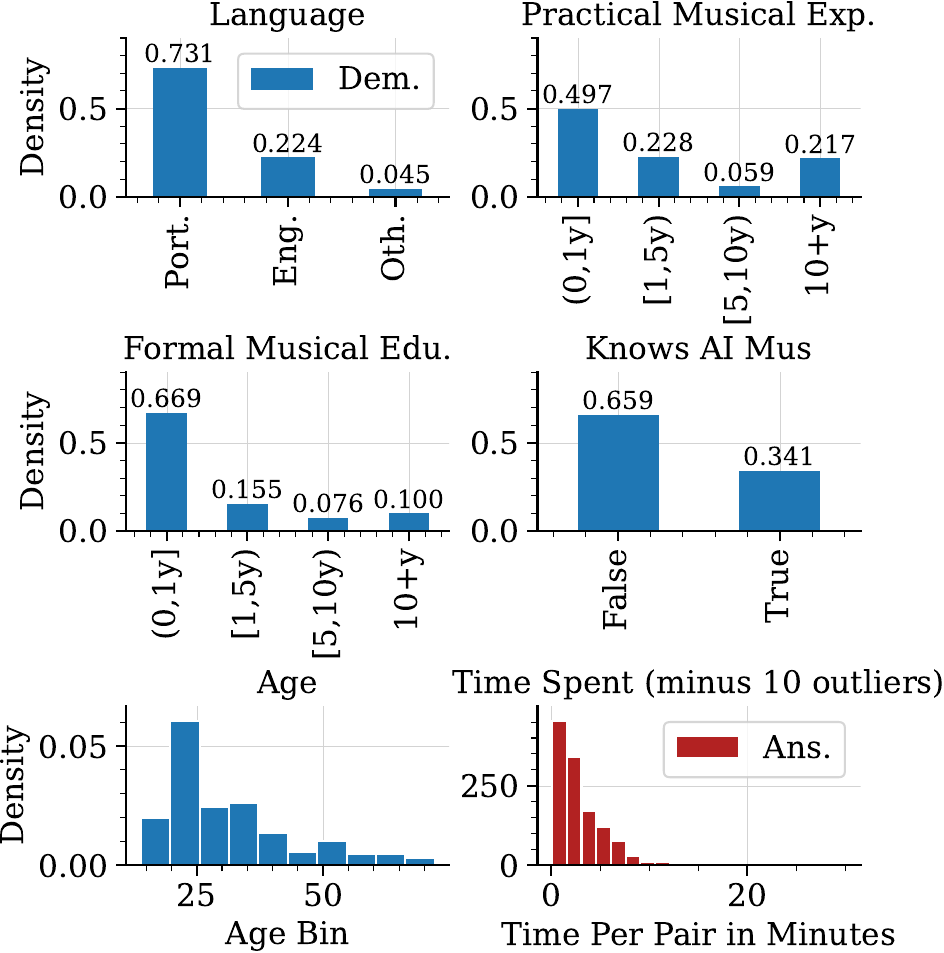}
        \captionof{figure}{Demographic and Answer Variables}
        \label{fig:demo}
    \end{minipage}
    \hfill 
    \begin{minipage}{0.48\textwidth}
        \scriptsize
        \centering
        \begin{tabular}{llrrc}
    \toprule
    &  & \textbf{Estimate} & \textbf{Pr(>$|z|$)} & \textbf{Sig.} \\
    \midrule
     & Intercept & -24.26 & 0.9968 & \\
    \midrule
    \rowcolor{SpringGreen} $\uparrow$ & Similar Pair & 0.61 & 0.0999 & * \\
    \midrule
    & Choice: Song A or B & 22.29 & 0.9971 & \\
    & Choice: Both Songs & 0.82 & 0.9999 & \\
    & Choice: Neither Song & 4.94 & 0.9994 & \\
    \rowcolor{SpringGreen} $\uparrow$ & $\log_{10}$(TimeSpent+1) & 0.49 & 0.0611 & * \\
    \midrule
    & Lang. Port. & -0.07 & 0.7621 & \\
    & Prac. Exp. 1 to 5 y & 0.42 & 0.1157 & \\
    \rowcolor{SpringGreen}  $\uparrow$ & Prac. Exp. 5 to 10 y & 0.92 & 0.0995 & * \\
    \rowcolor{SpringGreen}  $\uparrow$ & Prac. Exp. Over 10 y & 1.25 & 0.0009 & *** \\
    & Formal Edu. 1 to 5 y & -0.22 & 0.4803 & \\
    \rowcolor{Lavender} $\downarrow$ & Formal Edu. 5 to 10 y & -1.30 & 0.0086 & *** \\
    & Formal Edu. Over 10 y & -0.82 & 0.0614 & * \\
    \rowcolor{SpringGreen}  $\uparrow$ & Knowledge on AIM & 0.89 & 0.00005 & *** \\
    \rowcolor{Lavender} $\downarrow$ & Participants' Age & -0.03 & 0.0009 & *** \\
    \bottomrule
\end{tabular}
        \captionof{table}{Covariates *$p<.1$, **$<.05$, ***$<.01$}
        \label{tab:posterior_summary}
    \end{minipage}
\end{figure}

Out of these 308 participants, 290 performed the final demographic survey, summarized in Figure~\ref{fig:demo}. One can see that 73\% of participants were Portuguese speakers and 22\% were English speakers, whereas 5\% stated other native languages. Most participants do not have practical experience (50\%) nor formal musical education (67\%). Moreover, 34\% of participants have {\em some} knowledge of AIM. The mean age was 31 years (SD of 13), and the median was 27 years. The average time listeners spent on each pair was 2.98 minutes, excluding 10 outliers who had values over 30 minutes. 


\subsection{{\em When} listeners differentiate (RCCT analysis)}

In our RCCT, the following events needs to occur for a correct answer. A participant listens to a pair and decides that either song A or B is AIM (57\% of answers). Observe that the participant is always wrong when stating that both (20\%) or neither (16\%) pairs are AIM, or that they can't decide (7\%). Given our randomized trial, choices are mediated by the pair type ({\em similar} or {\em random}). Furthermore, the $\{A, B\}$ pair itself, the surveyed demographic traits of the listener, the time spent listening to the pair, the position of the pair (one to four), and other unseen traits of the participant may also affect the outcome. Next, we present results both controlling and not controlling for these endogenous factors.

We first compare listeners' success rates in distinguishing between song~A and song~B. With two choices at hand, a random guess yields an expected success rate of $\mathbb{E}[s] = 0.5$. Overall, across all pair types, the observed rate is $\hat{s}_o = 0.6$. For the \emph{random} set of pairs, $\hat{s}_r = 0.53$, which a binomial test shows is not significantly different from $0.5$ ($p > 0.05$). For the \emph{similar} set, $\hat{s}_s = 0.66$, significantly higher than both random guessing ($p < 10^{-9}$) and the overall rate ($p < 10^{-2}$). We further tested whether results change when restricting to songs with lyrics ($\hat{s}_r = 0.53$ with $p>0.05$, $\hat{s}_s = 0.75$ with $p<10^{-6}$) or excluding the \emph{classical} and \emph{ambient} genres, present only in the \emph{similar} set ($\hat{s}_s = 0.66; p>0.05$, $\hat{s}_r = 0.53; p<10^{-7}$). In both cases, the original findings held. Thus we have no reason to believe that results are confounded by genre or by the presence/absence of lyrics.


To study our control variables, we employed the Mixed Effects Logistic Model~\cite{Bates2010} described in Table~\ref{tab:posterior_summary}. This model was focused on predicting correct answers. Our model included pair-specific intercepts to capture differences in pair difficulty, as well as nested intercepts for participant ids and pair order. These control for the individual variability and ordering effects per participant. This model has a McFadden's $R^2=0.44$, which is more than acceptable~\cite{Mcfadden1973} for an explanatory model as ours. Nevertheless, in order to perform a sensitivity analysis, variations of this model were also considered (Appendix~\ref{appn:models}), as were Item-Response Theory~\cite{Baker2001} models (Appendix~\ref{appn:irf}). 

Using our model, we can see that Practical Experience (over five years) has a positive impact. This is also true for prior knowledge of AIM. Contrarily, age exhibited a negative association, with older participants being less likely to correctly identify AI‑generated songs. Surprisingly, a Formal Education of 5 to 10 years appears on the model with a negative effect. Nevertheless, we point out that there is a high correlation of Practical Experience and Formal Education. When we consider an alternative models without the Practical Experience factors, the effect disappears (Appendix~\ref{appn:models}).

These results indicate that the treatment effect of being exposed to a similar pair versus a random pair is of 13\% ($\hat{s}_s - \hat{s}_r = 0.13$). This effect is further mediated by characteristics of the participants. These findings provide evidence towards the {\em when} users differentiate. 

\subsection{{\em How} Listeners Differentiate (Mixed Methods Analysis of Feedback)}

We now turn our attention to the free-form textual answers. In total, we received 317 of such comments from 140 participants. To explore this data, we resorted to a manual qualitative coding. This coding was performed in {\em two} sessions, where in both of which three coders performed the coding. In each round, coders worked with a randomly sorted, per coder, list of answers. 

\begin{figure}[t!]
\begin{center}\includegraphics[width=1\linewidth]{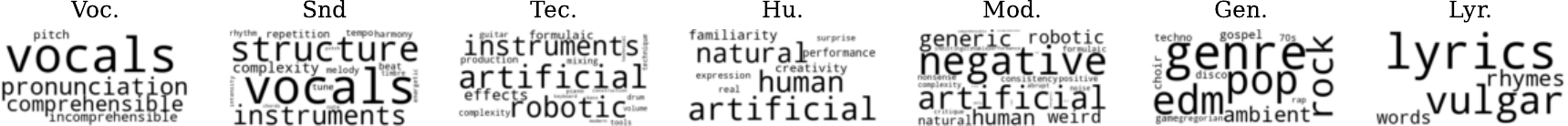}
\end{center}
\caption{Topics and tags. Word size is proportional to usage within topic. Top-7 overall frequency: vocals (369), lyrics (247), negative (231), artificial (224), generic (174), human (130), robotic (112)}
\label{f:topics}
\end{figure}

In the first round, the three coders tagged 100 answers with any textual labels they found useful over the course of a week. They then met for an hour to define seven major answer topics: \textit{vocals related, sound related, technical aspects, human aspects, modifiers, genre, and lyrics related}. For each topic, a set of non-exclusive tags was established (Figure~\ref{f:topics}). The second week-long coding round used these predefined tags. In this round, all answers were coded by all coders. Moreover, coders also noted when a tag applied to each song in the pair. Our quantitative analysis below focuses on answers with full coder agreement (3/3; 289 answers), while Appendix~\ref{appn:coding} qualitatively discusses these answers.

\begin{figure}[t!]
\begin{center}\includegraphics[width=.95\linewidth]{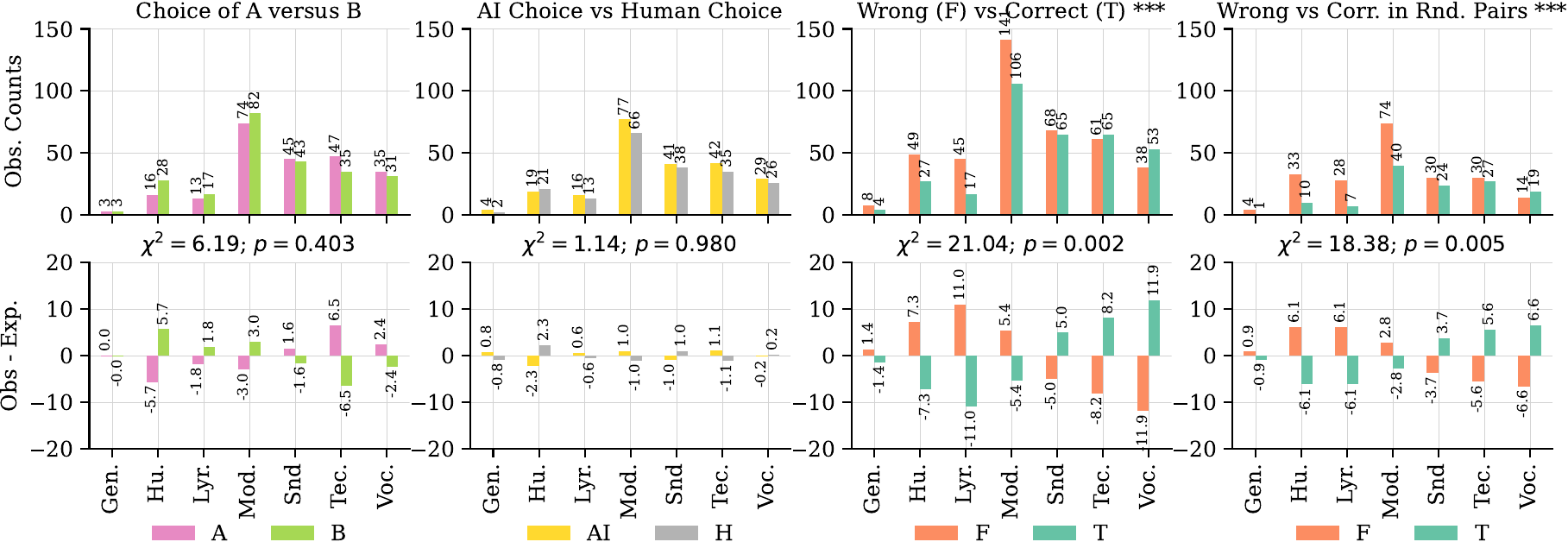}
\end{center}
\caption{Observed Topic Frequencies and Differences Towards the Expected. ***$p < .01$}
\label{f:chisq}
\end{figure}
 
We initially point out that, from the tags shown in Figure~\ref{f:topics}, participants commonly exploit vocal and lyrical aspects to justify their answers. 
 In order to unveil the impact of cues, in Figure~\ref{f:chisq} we consider the topic counts in different contexts. Firstly, we consider the difference in topic counts conditioned on  any choice, be it song (A) or (B), regardless of it being the correct choice. Secondly, we condition on the AIM versus human-made choice, again regardless of it being the correct choice or not. In our third context, we consider the condition of whether the answer was right or wrong. Lastly, we consider when the context of when the listener got the answer right for the more difficult {\em random} set. For each of these settings, we performed a $\chi^2$-test of differences  (also shown on the figure). 

Given that song order was random within a pair, there is no statistical difference in the usage of topics based on the choice of one pair over the other. This expected result serves as a comparison for the next ones. When we focus on the choice of AIM versus human-made, we again do not observe any differences. This indicates that when participants choose a song as AIM, the cues they employ for this belief are similar to those used to choose a human-made song. Nevertheless, when we consider correct versus incorrect choices, and correct versus incorrect choices in the harder random setting, statistical significance arises. In these cases, the usage of the contextually grounded cues, i.e., {\em sound, technical, and vocal} cues, help discern whether a piece of music is AIM.

These findings contribute to the understanding of human perception of AI-generated content. The cues listeners employ to distinguish AIM may also inform strategies for making AIM more indistinguishable (or deliberately distinguishable) from human creations. Conversely, we believe that our results may help guide the development of educational initiatives to help users recognize AIM.

\section{Conclusions} \label{s:conc}

The increasing usage of AI-generated content poses several challenges to the music industry. One of these challenges involves educating users to identify which content is AI-generated. Through a randomized controlled crossover trial and a mixed-methods coding study, we unveil both {\em when} listeners can identify AIM and {\em how} they currently perceive humanness versus non-humanness.

As future work, we aim to extend our study to music beyond the Western, educated, industrialized, rich, and democratic (WEIRD) domain from which our song samples originate. We also aim to develop similar initiatives for other media types (e.g., text, images, and video).

\bibliographystyle{abbrv}
\bibliography{bibs}

\begin{thebibliography}{10}

\bibitem{Alonso2020}
P.~Alonso-Jim{\'e}nez, D.~Bogdanov, J.~Pons, and X.~Serra.
\newblock Tensorflow audio models in {Essentia}.
\newblock In {\em Proc. ICASSP.}, 2020.

\bibitem{Baker2001}
F.~B. Baker.
\newblock {\em The basics of item response theory}.
\newblock ERIC, 2001.

\bibitem{Bates2010}
D.~M. Bates.
\newblock lme4: Mixed-effects modeling with r, 2010.

\bibitem{Bogdanov2019}
D.~Bogdanov, M.~Won, P.~Tovstogan, A.~Porter, and X.~Serra.
\newblock The mtg-jamendo dataset for automatic music tagging.
\newblock In {\em Proc. ML4MD at ICML.}, 2019.

\bibitem{Atlantic2025}
I.~Bogost.
\newblock Spotify’s top band is an ai mystery, 2025.
\newblock The Atlantic. Last Accessed on July 2025. \url{https://theatlantic.com/technology/archive/2025/07/velvet-sundown-ai-band-spotify/683410/}.

\bibitem{Candello2017}
H.~Candello, C.~Pinhanez, and F.~Figueiredo.
\newblock Typefaces and the perception of humanness in natural language chatbots.
\newblock In {\em Proc. CHI}, 2017.

\bibitem{Cetinic2022}
E.~Cetinic and J.~She.
\newblock Understanding and creating art with ai: Review and outlook.
\newblock {\em ACM transactions on multimedia computing, communications, and applications (TOMM)}, 18(2), 2022.

\bibitem{Chen2024}
R.~J. Chen, T.~Ding, M.~Y. Lu, D.~F. Williamson, G.~Jaume, A.~H. Song, B.~Chen, A.~Zhang, D.~Shao, M.~Shaban, et~al.
\newblock Towards a general-purpose foundation model for computational pathology.
\newblock {\em Nature medicine}, 30(3), 2024.

\bibitem{copet2023simple}
J.~Copet, F.~Kreuk, I.~Gat, T.~Remez, D.~Kant, G.~Synnaeve, Y.~Adi, and A.~D{\'e}fossez.
\newblock Simple and controllable music generation.
\newblock {\em Proc. NeuRIPs}, 2023.

\bibitem{Corbin1990}
J.~M. Corbin and A.~Strauss.
\newblock Grounded theory research: Procedures, canons, and evaluative criteria.
\newblock {\em Qualitative sociology}, 13(1), 1990.

\bibitem{Donahue2019}
C.~Donahue, H.~H. Mao, Y.~E. Li, G.~W. Cottrell, and J.~McAuley.
\newblock Lakhnes: Improving multi-instrumental music generation with cross-domain pre-training.
\newblock In {\em Proc. ISMIR.}, 2019.

\bibitem{Doyle2009}
L.~Doyle, A.-M. Brady, and G.~Byrne.
\newblock An overview of mixed methods research.
\newblock {\em Journal of research in nursing}, 14(2), 2009.

\bibitem{Musically2025}
S.~Dredge.
\newblock 10\% of new tracks uploaded to deezer are ai-generated music, 2025.
\newblock Musically. Last Accessed on July 2025. \url{https://musically.com/2025/01/27/10-of-new-tracks-uploaded-to-deezer-are-ai-generated-music/}.

\bibitem{Elgammal2017}
A.~Elgammal, B.~Liu, M.~Elhoseiny, and M.~Mazzone.
\newblock {CAN}: Creative adversarial networks generating “art” by learning about styles and deviating from style norms.
\newblock In {\em Proc. ICCC.}, 2017.

\bibitem{Epstein2023}
Z.~Epstein, A.~Hertzmann, I.~of~Human~Creativity, M.~Akten, H.~Farid, J.~Fjeld, M.~R. Frank, M.~Groh, L.~Herman, N.~Leach, et~al.
\newblock Art and the science of generative ai.
\newblock {\em Science}, 380(6650), 2023.

\bibitem{Grotschla2025}
F.~Gr{\"o}tschla, A.~Solak, L.~A. Lanzend{\"o}rfer, and R.~Wattenhofer.
\newblock Benchmarking music generation models and metrics via human preference studies.
\newblock In {\em Proc. ICASSP}. IEEE, 2025.

\bibitem{He2015}
K.~He, X.~Zhang, S.~Ren, and J.~Sun.
\newblock Delving deep into rectifiers: Surpassing human-level performance on imagenet classification.
\newblock In {\em Proc. CVPR.}, 2015.

\bibitem{Hernandez2022}
C.~Hernandez-Olivan, J.~A. Puyuelo, and J.~R. Beltran.
\newblock Subjective evaluation of deep learning models for symbolic music composition.
\newblock In {\em Proc. CHI Workshop on Generative AI and HCI}, 2022.

\bibitem{Ji2023}
S.~Ji, X.~Yang, and J.~Luo.
\newblock A survey on deep learning for symbolic music generation: Representations, algorithms, evaluations, and challenges.
\newblock {\em ACM Computing Surveys}, 56(1), 2023.

\bibitem{Jones2003}
B.~Jones and M.~G. Kenward.
\newblock {\em Design and analysis of cross-over trials}.
\newblock Chapman and Hall/CRC, 2003.

\bibitem{Musictech2025}
C.~Koe.
\newblock Study finds widespread ai use among music producers, 2025.
\newblock Musictech. Last Accessed on July 2025. \url{https://musictech.com/news/industry/study-widespread-ai-use-producers/}.

\bibitem{Kobis2021}
N.~Köbis and L.~D. Mossink.
\newblock Artificial intelligence versus maya angelou: Experimental evidence that people cannot differentiate ai-generated from human-written poetry.
\newblock {\em Computers in Human Behavior}, 114:106553, 2021.

\bibitem{Le2025}
D.-V.-T. Le, L.~Bigo, D.~Herremans, and M.~Keller.
\newblock Natural language processing methods for symbolic music generation and information retrieval: A survey.
\newblock {\em ACM Computing Surveys}, 57(7), 2025.

\bibitem{Lecamwasam2025}
K.~Lecamwasam and T.~R. Chaudhuri.
\newblock Exploring listeners' perceptions of ai-generated and human-composed music for functional emotional applications.
\newblock {\em arXiv preprint arXiv:2506.02856}, 2025.

\bibitem{Lerch2025}
A.~Lerch, C.~Arthur, N.~Bryan-Kinns, C.~Ford, Q.~Sun, and A.~Vinay.
\newblock Survey on the evaluation of generative models in music.
\newblock {\em arXiv preprint arXiv:2506.05104}, 2025.

\bibitem{Mcfadden1973}
D.~McFadden.
\newblock Conditional logit analysis of qualitative choice behavior.
\newblock {\em Frontier in Econometrics}, 1973.

\bibitem{Noll1966}
A.~M. Noll.
\newblock Human or machine: A subjective comparison of piet mondrian’s “composition with lines” (1917) and a computer-generated picture.
\newblock {\em The Psychological Record}, 16(1), 1966.

\bibitem{Ragot2020}
M.~Ragot, N.~Martin, and S.~Cojean.
\newblock Ai-generated vs. human artworks. a perception bias towards artificial intelligence?
\newblock In {\em Proc. CHI. Extended Abstracts}, 2020.

\bibitem{Russakovsky2015}
O.~Russakovsky, J.~Deng, H.~Su, J.~Krause, S.~Satheesh, S.~Ma, Z.~Huang, A.~Karpathy, A.~Khosla, M.~Bernstein, et~al.
\newblock Imagenet large scale visual recognition challenge.
\newblock {\em International journal of computer vision}, 115(3), 2015.

\bibitem{Santy2025}
S.~Santy, P.~Bhattacharya, M.~H. Ribeiro, K.~Allen, and S.~Oh.
\newblock When incentives backfire, data stops being human.
\newblock In {\em Proc. ICML.}, 2025.

\bibitem{Sarmento2024}
P.~Sarmento, J.~Loth, and M.~Barthet.
\newblock Between the ai and me: Analysing listeners' perspectives on ai-and human-composed progressive metal music.
\newblock In {\em Proc. ISMIR.}, 2024.

\bibitem{Silver2016}
D.~Silver, A.~Huang, C.~J. Maddison, A.~Guez, L.~Sifre, G.~Van Den~Driessche, J.~Schrittwieser, I.~Antonoglou, V.~Panneershelvam, M.~Lanctot, et~al.
\newblock Mastering the game of go with deep neural networks and tree search.
\newblock {\em Nature}, 529(7587), 2016.

\bibitem{Silver2018}
D.~Silver, T.~Hubert, J.~Schrittwieser, I.~Antonoglou, M.~Lai, A.~Guez, M.~Lanctot, L.~Sifre, D.~Kumaran, T.~Graepel, et~al.
\newblock A general reinforcement learning algorithm that masters chess, shogi, and go through self-play.
\newblock {\em Science}, 362(6419), 2018.

\bibitem{Turing1950}
A.~M. Turing.
\newblock Computing machinery and intelligence.
\newblock {\em Mind}, 59(236), 1950.

\bibitem{White2025}
C.~W. White, K.~Kapoor, N.~Cosme-Clifford, J.~Symons, and L.~von Mutius.
\newblock Humans perceive ai-generated music as less expressive than comparable human-made content.
\newblock {\em Available at SSRN 5087035}, 2025.

\bibitem{Wu2024}
Y.~Wu, K.~Chen, T.~Zhang, Y.~Hui, T.~Berg-Kirkpatrick, and S.~Dubnov.
\newblock Large-scale contrastive language-audio pretraining with feature fusion and keyword-to-caption augmentation.
\newblock In {\em Proc. ICASSP}, pages 1--5, 2023.

\bibitem{Xu2024}
A.~Xu, S.~Fang, H.~Yang, S.~Hosio, and K.~Yatani.
\newblock Examining human perception of generative content replacement in image privacy protection.
\newblock In {\em Proc. CHI}, 2024.

\bibitem{Yang2020}
L.-C. Yang and A.~Lerch.
\newblock On the evaluation of generative models in music.
\newblock {\em Neural Computing and Applications}, 32(9), 2020.

\end{thebibliography}

\pagebreak
\appendix

\section{Ethical Statement} \label{appn:ethics}

We observed that email addresses were used to track responses but were stored using a one-way, cryptographically secure hash algorithm. Participants were also given the option to provide limited demographic information, such as age and musical background. This procedure makes it virtually impossible to identify individual participants. As such, our study complies with the ethical guidelines and approval process of the Ethics Committee at the Federal University of Minas Gerais (UFMG), which stipulates that survey-based research must ensure that participants remain unidentifiable.

\section{Limitations} \label{appn:limit}

In our study, we had no control over the prompts used to generate AIM songs. We also had no control on how Jamendo songs were recorded and/or mixed. Due to these factors, we cannot control for prompt quality and recording quality, factors that may affect our results. Nevertheless, we do point out that we explored several pairs of songs to alleviate these possible effects. Moreover, we made use of the highest quality audio available for download for AIM (48kHz, 192kbps, stereo), as well as available on the MTG-Jamendo (44.1kHz, 320kbps, stereo) dataset. In order to reduce any biases, we also decided not to re-encode our audio files in any manner. That is, we make us of them as they are availale on AIM websites or on the MTG-Jamendo dataset. We further point that we have no reason to believe that the small difference in bit-rates across datasets has any impact on our results. That is, for the random experiment, participants were no better than random chance in determining AIM.

\section{An Item Response Theory Approach} \label{appn:irf}

As a second means to evaluate the research hypotheses, we also employed a IRT based psychometric approach. That is, to initially assess the psychometric quality of our instrument, we started with a 2-Parameter Logistic Model from Item Response Theory (IRT-2PL), i.e.:

\[
P(Y_{ij} = 1 \mid \theta_j) =
\frac{1}{1 + \exp\left[ -a_i \left( \theta_j - b_i \right) \right]}
\]

In this model, $Y_{ij}$ indicates the probability of participant $j$ answering item $i$ correctly.
\(a_i\) represents the discrimination parameter of item \(i\), indicating how sharply the probability of a correct response changes with ability. 
\(b_i\) is the difficulty parameter of item \(i\), corresponding to the ability level at which the probability of a correct response is 50\%. 
\(\theta_j\) denotes the latent ability of person \(j\), representing their position on the underlying ability scale. 

Observe that an IRT model is essentially a variation of a logistic model. For such a reason, we expect the similar results as in our main text. Moreover, and as is common on IRT analyzes, when summarizing parameter effects we rely on average effects (i.e., the average of all $\alpha_i$). 

From the perspective of the model above, the empirical difficulty of the items was notably high, as participants, on average, correctly answered only 34\% of the experimental questions. This difficulty was reflected in the IRT model with extremely low discrimination parameters ($a$), with a mean of only 0.016 for random pairs and 0.024 for similar pairs. According to established guidelines, discrimination values below 0.35 are considered ``very low'' or indicative of no effective discrimination \cite{Baker2001}. This result in line with our main text that found that overall, a moderate 10\%, (see the discussion on $\hat{s}_o = 0.6$ in Section~\ref{s:results}) increase from random guessing ($\mathbb{E}[s] = 0.5$) exists {\em overall}.


Despite the low overall discrimination, the experimental manipulation of presenting \textit{similar} music pairs had a statistically significant and large-magnitude effect on item quality. A formal comparison between the discrimination parameters of the two experimental groups is shown in Table~\ref{tab:irt}. A $t$-test result comparing these parameter is significant ($p < .001$) and the large effect size ($d = 2.84$) robustly confirm that the similarity context causally improves item quality.

\begin{table}[t!]
\centering
\begin{tabular}{lcccc}
\toprule
Experimental Group & Mean Discrimination & t-statistic & p-value & Cohen's $d$ \\
\midrule
Similar & 0.0238 & 6.856 & $< .001$ & 2.84 \\
Random & 0.0162 & & & \\
\bottomrule
\end{tabular}
\caption{Hypothesis Test for the Difference in Item Discrimination}
\label{tab:irt}
\end{table}

Overall, the IRT model has a McFadden's $R^2 \approx 0.003$. In this sense, due the inherent difficulty of the task, reflected in the extremely low discrimination parameters. This finding motivated the transition to methods from Item Response Theory to Mixed Methods models (as in our main text), which are less dependent on assumptions unsupported by the data.

Nevertheless, the IRT estimates still comply with our main findings on the causal effect of similar versus random pairs of items. In Appendix~\ref{appn:models}, we present different variations of our main-text model that also reach similar conclusions.

\section{Variations of the Logistic Model} \label{appn:models}

In the seven sub-tables shown in Table~\ref{tab:alt}, we present different variations of the model. In particular, we start with the model used on our main text. We here refer to the pairwise intercepts, as well as the nested participant by pair order intercepts as hierarchical variables. In this sense, the model used in our paper is described as the Full Hierarchical Model. 

In order to perform a sensitivity analysis, we  progressively remove covariates starting from these hierarchical effects (leading to the Full Non Hierarchical model). After we remove hierarchical effects, we add a single new variable that accounts for pair order (this was a nested effect in our full model). Next, we remove major groups of endogenous traits until we reach a model with no participant control variables at all.

On all variations of the models, the pair type (similar versus random) was a significant effect. Also observe that the negative effect of Formal Education from 5 to 10 years, observed in the full model used in the paper, also disappears when we simplify the model. This is likely due to the correlation of Formal Education and Practical Experience. It is also interesting that when we remove Answer traits, the ordering of pairs becomes significant and negative. 
Overall, this negative value {\em may} indicate a trend that participants become more tired over time. Thus, they are more likely to make mistakes.

\begin{figure}[p]
    \centering
    \begin{minipage}{0.48\textwidth}
    \captionof*{table}{Full Hierarchical Model (Paper model).}
\scriptsize
        \centering
        \begin{tabular}{llrrc}
            \toprule
            & & \textbf{Estimate} & \textbf{Pr(>$|z|$)} & \textbf{Sig.} \\
            \midrule
             & Intercept & -24.26 & 0.9968 & \\
            \midrule
            \rowcolor{SpringGreen} $\uparrow$ & Similar Pair & 0.61 & 0.0999 & * \\
            \midrule
            & Choice: Song A or B & 22.29 & 0.9971 & \\
            & Choice: Both Songs & 0.82 & 0.9999 & \\
            & Choice: Neither Song & 4.94 & 0.9994 & \\
            \rowcolor{SpringGreen} $\uparrow$ & $\log_{10}$(TimeSpent+1) & 0.49 & 0.0611 & * \\
            \midrule
            & Lang. Port. & -0.07 & 0.7621 & \\
            & Prac. Exp. 1 to 5 y & 0.42 & 0.1157 & \\
            \rowcolor{SpringGreen}  $\uparrow$ & Prac. Exp. 5 to 10 y & 0.92 & 0.0995 & * \\
            \rowcolor{SpringGreen}  $\uparrow$ & Prac. Exp. Over 10 y & 1.25 & 0.0009 & *** \\
            & Formal Edu. 1 to 5 y & -0.22 & 0.4803 & \\
            \rowcolor{Lavender} $\downarrow$ & Formal Edu. 5 to 10 y & -1.30 & 0.0086 & *** \\
            & Formal Edu. Over 10 y & -0.82 & 0.0614 & * \\
            \rowcolor{SpringGreen}  $\uparrow$ & Knowledge on AIM & 0.89 & 0.00005 & *** \\
            \rowcolor{Lavender} $\downarrow$ & Participants' Age & -0.03 & 0.0009 & *** \\
            \bottomrule
        \end{tabular}
    \end{minipage}
    \hfill 
    \begin{minipage}{0.48\textwidth}
        \captionof*{table}{Full Non Hierarchical.}

        \scriptsize
        \centering
        \begin{tabular}{llrrc}
    \toprule
    & & \textbf{Estimate} & \textbf{Pr(>$|z|$)} & \textbf{Sig.} \\
    \midrule
     & Intercept & -21.38 & 0.9854 & \\
    \midrule
    \rowcolor{SpringGreen} $\uparrow$ & Similar Pair & 0.58 & 0.0005 & *** \\
    & Question Order & -0.02 & 0.8121 & \\
    \midrule
    & Choice: Song A or B & 19.95 & 0.9864 & \\
    & Choice: Both Songs & -0.13 & 0.9999 & \\
    & Choice: Neither Song & -0.27 & 0.9998 & \\
    \rowcolor{SpringGreen} $\uparrow$ & $\log_{10}$(TimeSpent+1) & 0.40 & 0.0863 & * \\
    \midrule
    & Lang. Port. & -0.12 & 0.5519 & \\
    & Prac. Exp. 1 to 5 y & 0.38 & 0.0936 & * \\
    & Prac. Exp. 5 to 10 y & 0.84 & 0.0769 & * \\
    \rowcolor{SpringGreen} $\uparrow$ & Prac. Exp. Over 10 y & 1.02 & 0.0016 & ** \\
    & Formal Edu. 1 to 5 y & -0.18 & 0.5037 & \\
    \rowcolor{Lavender} $\downarrow$ & Formal Edu. 5 to 10 y & -0.99 & 0.0173 & * \\
    & Formal Edu. Over 10 y & -0.63 & 0.0968 & * \\
    \rowcolor{SpringGreen} $\uparrow$ & Knowledge on AIM & 0.78 & 0.00003 & *** \\
    \rowcolor{Lavender} $\downarrow$ & Participants' Age & -0.03 & 0.00024 & *** \\
    \bottomrule
\end{tabular}
    \end{minipage}
    \vspace{2em}
    
    \begin{minipage}{0.48\textwidth}
    \captionof*{table}{Non-Hier. Without Practical Exp.}

\scriptsize
        \centering
        \begin{tabular}{llrrc}
    \toprule
    & & \textbf{Estimate} & \textbf{Pr(>$|z|$)} & \textbf{Sig.} \\
    \midrule
     & Intercept & -22.01 & 0.9851 & \\
    \midrule
    \rowcolor{SpringGreen} $\uparrow$ & Similar Pair & 0.58 & 0.0005 & *** \\
    & Question Order & 0.00 & 0.9864 & \\
    \midrule
    & Choice: Song A or B & 19.89 & 0.9865 & \\
    & Choice: Both Songs & -0.14 & 0.9999 & \\
    & Choice: Neither Song & -0.27 & 0.9998 & \\
    \rowcolor{SpringGreen} $\uparrow$ & $\log_{10}$(TimeSpent+1) & 0.55 & 0.0181 & * \\
    \midrule
    & Lang. Port. & -0.12 & 0.5645 & \\
    & Formal Edu. 1 to 5 y & 0.15 & 0.5256 & \\
    & Formal Edu. 5 to 10 y & -0.22 & 0.4574 & \\
    & Formal Edu. Over 10 y & 0.11 & 0.6996 & \\
    \rowcolor{SpringGreen} $\uparrow$ & Knowledge on AIM & 0.86 & 0.000003 & *** \\
    \rowcolor{Lavender} $\downarrow$ & Participants' Age & -0.03 & 0.0003 & *** \\
    \bottomrule
\end{tabular}
    \end{minipage}
    \hfill 
    \begin{minipage}{0.48\textwidth}
    \captionof*{table}{Non-Hier. Without Formal Exp.}

        \scriptsize
        \centering
\begin{tabular}{llrrc}
    \toprule
    & & \textbf{Estimate} & \textbf{Pr(>$|z|$)} & \textbf{Sig.} \\
    \midrule
     & Intercept & -21.91 & 0.9852 & \\
    \midrule
    \rowcolor{SpringGreen} $\uparrow$ & Similar Pair & 0.59 & 0.0004 & *** \\
    & Question Order & 0.00 & 0.9617 & \\
    \midrule
    & Choice: Song A or B & 19.85 & 0.9865 & \\
    & Choice: Both Songs & -0.17 & 0.9999 & \\
    & Choice: Neither Song & -0.30 & 0.9998 & \\
    \rowcolor{SpringGreen} $\uparrow$ & $\log_{10}$(TimeSpent+1) & 0.51 & 0.0291 & * \\
    \midrule
    & Lang. Port. & -0.12 & 0.5583 & \\
    & Prac. Exp. 1 to 5 y & 0.29 & 0.1679 & \\
    & Prac. Exp. 5 to 10 y & 0.19 & 0.5927 & \\
    \rowcolor{SpringGreen} $\uparrow$ & Prac. Exp. Over 10 y & 0.52 & 0.0209 & * \\
    \rowcolor{SpringGreen} $\uparrow$ & Knowledge on AIM & 0.77 & 0.000033 & *** \\
    \rowcolor{Lavender} $\downarrow$ & Participants' Age & -0.03 & 0.00035 & *** \\
    \bottomrule
\end{tabular}

    \end{minipage}
    \vspace{2em}

    \begin{minipage}{0.48\textwidth}
            \captionof*{table}{Non-Hier. Question/Answer Traits Only.}

\scriptsize
        \centering
        \begin{tabular}{llrrc}
    \toprule
    & & \textbf{Estimate} & \textbf{Pr(>$|z|$)} & \textbf{Sig.} \\
    \midrule
     & Intercept & -22.08 & 0.9853 & \\
    \midrule
    \rowcolor{SpringGreen} $\uparrow$ & Similar Pair & 0.55 & 0.0006 & *** \\
    & Question Order & 0.01 & 0.8837 & \\
    \midrule
    & Choice: Song A or B & 19.98 & 0.9867 & \\
    & Choice: Both Songs & -0.02 & 0.99999 & \\
    & Choice: Neither Song & -0.09 & 0.99995 & \\
    \rowcolor{SpringGreen} $\uparrow$ & Time Spent & 0.43 & 0.0466 & * \\
    \bottomrule
\end{tabular}
    \end{minipage}
    \hfill 
    \begin{minipage}{0.48\textwidth}
            \captionof*{table}{Non-Hier. Question/Demographic Traits Only.}
        \scriptsize
        \centering
\begin{tabular}{llrrc}
    \toprule
    & & \textbf{Estimate} & \textbf{Pr(>$|z|$)} & \textbf{Sig.} \\
    \midrule
     & Intercept & -0.25 & 0.3982 & \\
    \midrule
    \rowcolor{SpringGreen} $\uparrow$ & Similar Pair & 0.33 & 0.0105 & * \\
    \rowcolor{Lavender} $\downarrow$ & Question Order & -0.12 & 0.0296 & * \\
    \midrule
    & Lang. Port. & 0.16 & 0.3101 & \\
    & Formal Edu. 1 to 5 y & -0.19 & 0.3283 & \\
    & Formal Edu. 5 to 10 y & -0.32 & 0.2673 & \\
    & Formal Edu. Over 10 y & -0.19 & 0.4845 & \\
    \rowcolor{SpringGreen} $\uparrow$ & Prac. Exp. 1 to 5 y & 0.36 & 0.0347 & * \\
    & Prac. Exp. 5 to 10 y & 0.36 & 0.2441 & \\
    \rowcolor{SpringGreen} $\uparrow$ & Prac. Exp. Over 10 y & 0.64 & 0.0040 & ** \\
    \rowcolor{SpringGreen} $\uparrow$ & Knowledge on AIM & 0.55 & $\approx 0$ & *** \\
    \rowcolor{Lavender} $\downarrow$ & Participants' Age & -0.02 & $\approx 0$ & *** \\
    \bottomrule
\end{tabular}
    \end{minipage}
    \vspace{2em}
        
    \begin{minipage}{\textwidth}
    \captionof*{table}{No Participant Covariates}

    \scriptsize
    \centering
    \begin{tabular}{llrrc}
    \toprule
    & & \textbf{Estimate} & \textbf{Pr(>$|z|$)} & \textbf{Sig.} \\
    \midrule
    \rowcolor{Lavender} $\downarrow$ & Intercept & -0.51 & 0.01 & ** \\
    \midrule
    \rowcolor{SpringGreen} $\uparrow$ & Similar Pair & 0.31 & 0.012 & * \\
    \rowcolor{Lavender} $\downarrow$ & Question Order & -0.11 & 0.037 & * \\
    \bottomrule
    \end{tabular}
    \end{minipage}
    \vspace{1em}

    \captionof{table}{Alternative Models. Observe that we progressively remove covariates. Regardless of the model, the exposure to similar pairs is always significant.}
    \label{tab:alt}
\end{figure}

\section{Qualitative Analysis of Textual Feedback}
\label{appn:coding}


In this appendix, we present a brief qualitative exploration of some of the answers provided by participants. We group these answers by topics, that were defined on our main text.

We begin with an exploration of answers that were \textbf{Sound Related} and \textbf{Technical Aspects Related}. Regarding sound aspects, we observe that the listener's perception on the musical features plays an important role for the task of differentiating AIM from human-made. On the technical side, listeners presented their perception on the audio quality, the effects, the production and the tools used to create the track, from their point of view.

(translated) \textit{[..] In general, both songs may have been produced using samples, synthesizers, or software that wouldn’t be classified as AI but aren’t exactly human either.}.

\textit{Audio quality on track 2 makes it sound like it was generated}

\textit{Track 2 immediately seemed more human cause it made better use of stereo sound.}

\textbf{Vocals Related:} When considering vocal aspects, participants took into account cues such as pronunciation, technical performance, singing quality, as well as others. Some examples are:

\textit{"The first song felt awkward, the singer's voice failed sometimes. The second one was more smooth, I could hear the air in the microphone, in the p's and b's"}. Notice the focus on pronunciation

\textit{(translated) "The voices on the second track sounded a bit robotic to me, as if they were artificial, created by AI, and I couldn't pick up on any singing strategies (like head and chest voices) throughout the song."}. Here the focus is on singing performance. this is similar to the example below: 

\textit{"In the beginning, I guessed number 1 was AI-generated, but after listening number 2, where the voice is not fluid as expected, I decided that the track 2 is AI-generated, instead of 1."}.

Some cases focus on whether the voice was robotic or not:
\textit{"The singer's voice in track 2 sounds more robotic"}

Or if it sounds like a voice recorded live:
\textit{"Sounds like a live recording, especially because the lead vocal is quieter than it should be [...]."}.









\textbf{Human vs Artificial Aspects.} Listeners considered factors such as creativity, expression, performance, naturalness, in order to present human or artificial aspects of a specific track. Some comments brought their intuition on judging how "natural" a track seemed, with an attempt to justify this perception.

\textit{(translated) "AI-generated audio, from videos like those that have been trending, has a noise at the end of each word. And the first audio also has... That's why it's believed to be AI."}. Observe the focus on pronunciation. This is also an example of a feedback that focused on vocal aspects. 

\textit{[...] the singer’s voice sounds unnatural, recognizably unnatural, something that’s not quite humanly natural. [...] the \#2 track sounds ok, creative like humans would do.}. Focus on AI versus human creativity.

{\textit{[...] It's a stage feeling to describe but it sounds very human in the way it sound less like other human songs you hear normally. Also it sounds very good. The second it sound very artificial and in my opinion kinda uninspired too [...].} Focus on AI versus human performance.

\textit{The first song seems very Human made, the pronounciation was really human like. The vocal
intonations felt real. Also the deeper voice in some parts seems like a human touch. [...]}. Focus on AI versus human feeling.

 \textit{Some of the vocal resonances and the way the vocals were false made me think both tracks were artificially created}. Same as above.

\textbf{Lyrics Related:} Listeners heavily consider the lyric's quality when evaluating songs. Participants tend to classify as AI what they consider as incoherence in the text. 

\textit{"Track 2 lyrics don't make any sense but, the song sounds good. That is why I believe track 2 is AI generated."}. 

Additionally, the content of the lyrics is viewed as a tip for some participants. 

\textit{"The lyrics in track 1 were what made me think it was AI, they were a bit silly [...]"}. 

Also, the poetic structure of lyrics (rhymes) is pointed out as a characteristic that some users employ to differ artificial from human musics. 

\textit{"Track 1 is the first one I hear that I think current AI models couldn’t replicate. The flow and the rhymes are too complex to be an AI song [...]."}.

\textit{"[...] both lyrics sound like made by AI, like the more average pop songs ever created."}

\textit{"Track 1 is the first one I hear that I think current AI models couldn’t replicate. The flow and the rhymes are too complex to be an AI song (I’m not saying AI will never do such songs). I’m almost certain it’s human. [...]"}



\textbf{Genre Related:} Some users highlight the commonness of elements present in the song, correlating them with the usual aspects seen on frequently distributed musics of the genre. 

\textit{"Track 1 could be a solo for any 80's hard rock band [...]"}. 

In contrast, a number of listeners specify which of these elements they notice as \textbf{not} being common; observe that, in this case, usually the listener ranges at different topics to conclude about the origin of the music. 

\textit{(translated) "The instrumentation of the first song doesn't really match the theme, which seems to be somewhat gospel. [...]"}. Also, a few participants pointed out their previous familiarity on AIM models capacities to create convincing songs on various genres. 

\textit{"[...] I just think its easier for AI to make a rap song than a heavy metal song without messing up too much."}.

Furthermore, the match between lyrics and genre was also emphasized.

\textit{"The lyrics to genre match don't seem to make sense in the first track."}

\textbf{Modifiers}. Modifiers were used to provide adjectives to the coders. These could be used in conjunction with another tag or as to describe the feedback as well.

\textit{(translated) I think the first one is strange, usually rappers don't repeat the same thing so many times.} (repetitive)

\textit{Again, for me it comes to things being generic. I know there are some awfully written songs made by humans, but even then you can hear some distinctions in their voices in different moments (which is not the case on track 2).} (generic)

\textit{Something about Track 1 sounded too perfect, too contrived to be Human. I was pretty confident with saying Track 2 was humans, by all the layers sounds and dimensions it had.} (perfection)

\textit{The first has a weird noise on the voice. The second the vocalist seams that don't breath, is odd} (weirdness)

\textit{I hope track 1 is AI generated, since it  sounds idiotic. Track 2 lyrics sounds like something that AI could create trying to emulate rap music} (critique)

\textit{The first one seems to have nonsensical lyrics, and the second one seems to have strange lyrics as well.} (nonsense)

\textit{[...]  Nonetheless, they were both very good recordings and hard to tell the difference.} (positive)

\newpage

\section*{NeurIPS Paper Checklist}

\begin{enumerate}

\item {\bf Claims}
    \item[] Question: Do the main claims made in the abstract and introduction accurately reflect the paper's contributions and scope?
    \item[] Answer: \answerYes{} 
    \item[] Justification: Claims made in Section~\ref{s:intro} are based on the empirical results of Section~\ref{s:results}. 

\item {\bf Limitations}
    \item[] Question: Does the paper discuss the limitations of the work performed by the authors?
    \item[] Answer: \answerYes{} 
    \item[] Justification: See Appendix~\ref{appn:limit}

\item {\bf Theory assumptions and proofs}
    \item[] Question: For each theoretical result, does the paper provide the full set of assumptions and a complete (and correct) proof?
    \item[] Answer: \answerNA{} 
    \item[] Justification: This is a mixed-methods user study, we do not present any proofs.

    \item {\bf Experimental result reproducibility}
    \item[] Question: Does the paper fully disclose all the information needed to reproduce the main experimental results of the paper to the extent that it affects the main claims and/or conclusions of the paper (regardless of whether the code and data are provided or not)?
    \item[] Answer: \answerYes{} 
    \item[] Justification: Our experimental setup is described in details on Section~\ref{s:mm}. 

\item {\bf Open access to data and code}
    \item[] Question: Does the paper provide open access to the data and code, with sufficient instructions to faithfully reproduce the main experimental results, as described in supplemental material?
    \item[] Answer: \answerYes{} 
    \item[] Justification: See Section~\ref{s:mm} for our links to the study platform, the data collected, and our source code.

\item {\bf Experimental setting/details}
    \item[] Question: Does the paper specify all the training and test details (e.g., data splits, hyperparameters, how they were chosen, type of optimizer, etc.) necessary to understand the results?
    \item[] Answer: \answerNA{} 
    \item[] Justification: This is a mixed-methods user study that relies on AI generated data from commercial models. We did not train any model.

\item {\bf Experiment statistical significance}
    \item[] Question: Does the paper report error bars suitably and correctly defined or other appropriate information about the statistical significance of the experiments?
    \item[] Answer: \answerYes{} 
    \item[] Justification: All of our quantitative findings are backed by statistical tests. See Section~\ref{s:results}, Appendix~\ref{appn:irf}, and Appendix~\ref{appn:models}.

\item {\bf Experiments compute resources}
    \item[] Question: For each experiment, does the paper provide sufficient information on the computer resources (type of compute workers, memory, time of execution) needed to reproduce the experiments?
    \item[] Answer: \answerNA{} 
    \item[] Justification: We did not perform traditional ML experiments.
    
\item {\bf Code of ethics}
    \item[] Question: Does the research conducted in the paper conform, in every respect, with the NeurIPS Code of Ethics \url{https://neurips.cc/public/EthicsGuidelines}?
    \item[] Answer: \answerYes{} 

\item {\bf Broader impacts}
    \item[] Question: Does the paper discuss both potential positive societal impacts and negative societal impacts of the work performed?
    \item[] Answer: \answerYes{} 
    \item[] Justification: Our main motivation, from the abstract, is on increasing the digital literacy on AI-generated content. 

\item {\bf Safeguards}
    \item[] Question: Does the paper describe safeguards that have been put in place for responsible release of data or models that have a high risk for misuse (e.g., pretrained language models, image generators, or scraped datasets)?
    \item[] Answer: \answerNA{} 
    \item[] Justification: We do not propose novel methods. We only release result data from our study. 
    

\item {\bf Licenses for existing assets}
    \item[] Question: Are the creators or original owners of assets (e.g., code, data, models), used in the paper, properly credited and are the license and terms of use explicitly mentioned and properly respected?
    \item[] Answer: \answerYes{} 
    \item[] Justification: We scrapped public Reddit, Suno and YouTube data but do not release it. 

\item {\bf New assets}
    \item[] Question: Are new assets introduced in the paper well documented and is the documentation provided alongside the assets?
    \item[] Answer: \answerYes{} 
    \item[] Justification: In our results link, our analysis notebooks are commented. We also provide an explanation of our result data.
    

\item {\bf Crowdsourcing and research with human subjects}
    \item[] Question: For crowdsourcing experiments and research with human subjects, does the paper include the full text of instructions given to participants and screenshots, if applicable, as well as details about compensation (if any)? 
    \item[] Answer: \answerYes{} 
    \item[] Justification: We provide the link to our study website on the main text.

\item {\bf Institutional review board (IRB) approvals or equivalent for research with human subjects}
    \item[] Question: Does the paper describe potential risks incurred by study participants, whether such risks were disclosed to the subjects, and whether Institutional Review Board (IRB) approvals (or an equivalent approval/review based on the requirements of your country or institution) were obtained?
    \item[] Answer: \answerYes{} 
    \item[] Justification: See Appendix~\ref{appn:ethics}.
    

\item {\bf Declaration of LLM usage}
    \item[] Question: Does the paper describe the usage of LLMs if it is an important, original, or non-standard component of the core methods in this research? Note that if the LLM is used only for writing, editing, or formatting purposes and does not impact the core methodology, scientific rigorousness, or originality of the research, declaration is not required.
    \item[] Answer: \answerYes{} 
    \item[] Justification: LLMs were used for grammar checks and translations. Not for generating original content.
    

\end{enumerate}

\end{document}